%% file: dapper_2018_arxiv.tex
\documentclass[conference]{IEEEtran}
\IEEEoverridecommandlockouts
\usepackage{cite}
\usepackage{amsmath,amssymb,amsfonts}
\usepackage{algorithmic}
\usepackage{graphicx}
\usepackage{textcomp}
\usepackage{xcolor}
\usepackage{booktabs}
\usepackage{tikz}
\def\BibTeX{{\rm B\kern-.05em{\sc i\kern-.025em b}\kern-.08em
    T\kern-.1667em\lower.7ex\hbox{E}\kern-.125emX}}

\input{equation_shorthands}

\newcommand\copyrighttext{%
  \footnotesize \textcopyright 2018 IEEE. Personal use of this material is permitted.
  Permission from IEEE must be obtained for all other uses, in any current or future
  media, including reprinting/republishing this material for advertising or promotional
  purposes, creating new collective works, for resale or redistribution to servers or
  lists, or reuse of any copyrighted component of this work in other works.}
\newcommand\copyrightnotice{%
\begin{tikzpicture}[remember picture,overlay]
\node[anchor=south,yshift=10pt] at (current page.south){\fbox{\parbox{\dimexpr\textwidth-\fboxsep-\fboxrule\relax}{\copyrighttext}}};
\end{tikzpicture}%
}

\begin{document}

\title{DAPPER: Scaling Dynamic Author Persona Topic Model to Billion Word Corpora}

\author{\IEEEauthorblockN{Robert Giaquinto}
\IEEEauthorblockA{\textit{Dept. of Computer Science and Engineering} \\
\textit{University of Minnesota}\\
Twin Cities, USA \\
giaquinto.ra@gmail.com}
\and
\IEEEauthorblockN{Arindam Banerjee}
\IEEEauthorblockA{\textit{Dept. of Computer Science and Engineering} \\
\textit{University of Minnesota}\\
Twin Cities, USA \\
banerjee@cs.umn.edu}
}

\maketitle
\copyrightnotice

\begin{abstract}
Extracting common narratives from multi-author dynamic text corpora requires complex models, such as the Dynamic Author Persona (DAP) topic model. However, such models are complex and can struggle to scale to large corpora, often because of challenging non-conjugate terms. To overcome such challenges, in this paper we adapt new ideas in approximate inference to the DAP model, resulting in the DAP Performed Exceedingly Rapidly (DAPPER) topic model. Specifically, we develop Conjugate-Computation Variational Inference (CVI) based variational Expectation-Maximization (EM) for learning the model, yielding fast, closed form updates for each document, replacing iterative optimization in earlier work. Our results show significant improvements in model fit and training time without needing to compromise the model's temporal structure or the application of Regularized Variation Inference (RVI). We demonstrate the scalability and effectiveness of the DAPPER model by extracting health journeys from the CaringBridge corpus --- a collection of 9 million journals written by 200,000 authors during health crises.
\end{abstract}


\input{1_introduction}
\input{2_background}

\input{3_dap}

\input{4_cvi_updates}

\input{5_experiments}

\input{6_results}

\input{7_conclusion}

\input{8_acknowledgements}

\bibliographystyle{IEEEtran}
\bibliography{dapper_2018_arxiv}

\end{document}

%% file: equation_shorthands.tex
\DeclareMathOperator*{\argmin}{arg\,min}
\DeclareMathOperator*{\argmax}{arg\,max}
\DeclareMathOperator{\Tr}{Tr}

\newcommand{\wt}{\widetilde}
\newcommand{\lambdab}{\pmb{\lambda}}
\newcommand{\Lambdab}{\pmb{\Lambda}}
\newcommand{\betab}{\pmb{\beta}}
\newcommand{\mub}{\pmb{\mu}}
\newcommand{\phib}{\pmb{\phi}}
\newcommand{\Phib}{\pmb{\Phi}}

\newcommand{\etab}{\pmb{\eta}}
\newcommand{\taub}{\pmb{\tau}}
\newcommand{\thetab}{\pmb{\theta}}
\newcommand{\Thetab}{\pmb{\Theta}}
\newcommand{\alphab}{\pmb{\alpha}}
\newcommand{\kappab}{\pmb{\kappa}}
\newcommand{\Sigmab}{\pmb{\Sigma}}
\newcommand{\gammab}{\pmb{\gamma}}
\newcommand{\deltab}{\pmb{\delta}}
\newcommand{\m}{\mathbf{m}}
\newcommand{\z}{\mathbf{z}}
\newcommand{\x}{\mathbf{x}}

\newcommand{\w}{\mathbf{w}}
\newcommand{\y}{\mathbf{y}}
\newcommand{\E}{\mathbb{E}}
\DeclareMathOperator{\diag}{diag}

%% file: 1_introduction.tex
\section{Introduction}
Topic modeling is a popular technique for automatically discovering compact, interpretable, latent representations of corpora. Many corpora exhibit an important structure, such as authorship or a temporal dependency between documents. Classic topic models like Latent Dirichlet Allocation (LDA) scale to large datasets \cite{Blei2003,Hoffman2010,Hoffman2013}, but do not account for any special structure in the corpus. Subsequent topic models are designed around such corpora, or are reparameterized to capture other features in the texts. For instance, the Correlated Topic Model (CTM) captures correlations between topics \cite{Lafferty2006}. The added complexity of these models comes at a cost, however. In the case of CTM, the model is parameterized with non-conjugate terms --- resulting in an additional variational parameter and requiring conjugate gradient descent to be run repeatedly on each document of the corpus. Up until recently CTM defined the standard approach in dealing with non-conjugate terms in variational inference. Topic models uniquely designed for corpora with a temporal structure, such as Dynamic Topic Model (DTM) and Continuous Time Dynamic Topic Model (CDTM), face similar issues as the CTM \cite{Blei2006a,Wang2008}. In each of these models the scalability is compromised by non-conjugate terms.

In recent work, the Dynamic Author-Persona (DAP) topic model was introduced for corpora with multiple authors writing over time \cite{Giaquinto2018}. DAP represents each author by a latent persona --- where personas capture the propensity to discuss certain topics over time. However, inference in DAP inherited the challenges with non-conjugacy from CTM and DTM.

In this paper, we seek to improve the scalability of the DAP topic model. Our approach is to adapt new ideas in approximate inference to DAP's variational Expectation-Maximization (EM) algorithm. Specifically, we develop a Conjugate-Computation Variation Inference (CVI) based variational EM algorithm, a powerful approach for transforming inference in non-conjugate models to conjugate models, leading to fast, closed form updates to parameters \cite{Khan2017}. The advantage of CVI over other related approaches is that it preserves the closed form updates to parameters in the conjugate terms. We show how a CVI based inference algorithm applies to a complex, temporal topic model like DAP, and how this new inference algorithm improves model performance and dramatically reduces the time required to train the model.

Our primary motivation for developing a faster inference algorithm for the DAP model is the desire to scale the model to the CaringBridge (CB) corpus, which is a collection of 9 million journals ($\approx$1 billion words) written by approximately 200,000 authors during a health crisis. CaringBridge journals are written by patients and caregivers and posted to the CaringBridge website, to be shared privately with friends and family. The CB corpus holds enormous potential for insights on the challenges and experiences faced by those with serious, and often life threatening illnesses. The size and complexity of the data, however, present a modeling challenge too great, until now.

Our results show that the DAPPER model achieves likelihoods better than competing models, including LDA, DTM, CDTM, and DAP. Moreover, we show that DAPPER's conjugate-computation updates result in significant improvements in speed over its predecessor. Finally, we demonstrate the scalability of the DAPPER model by training it to the CB and Signal Media One-Million News Article corpora, and share the compelling narratives found by DAPPER's latent personas.

The rest of the paper is as follows: in Section \ref{background}, a background on recent advances in approximate inference is given. Section \ref{model} presents a brief overview of the DAP model. Section \ref{cvi} details the CVI approach for accelerating the DAP model and describes a connection between CVI and expectation propagation. Section \ref{experiments} introduces the evaluation datasets and procedures. Section \ref{results} shares the results of the experiments. Finally, Section \ref{conclusion} summarizes the contributions of this paper.

%% file: 2_background.tex
\section{Background}
\label{background}
Approximate inference plays an important role in fitting complex probabilistic graphical models (PGM) which often have intractable posteriors and cannot be computed exactly. Interest in approximate inference techniques like variational inference, is growing because it tends to scale better than classical techniques, such as Markov Chain Monte Carlo \cite{blei2017variational}. Variational inference, in particular, transforms the inference problem into an optimization problem with the goal of finding hidden variables $\z$ to the variational distribution $q$ such that $q(\z)$ closely approximates the posterior $p(\z \mid \y)$, where $\y$ are the observed data \cite{Jordan1999}. This equates to minimizing KL divergence between the approximate and true posterior:

%\small
\begin{equation*}
q^*(\z) = \argmin_{q \in \mathcal{Q}} KL( q(\z) \mid \mid p(\z \mid \y))~.
\end{equation*}
%\normalsize

In PGMs the dependency between nodes, and their corresponding probability distributions, can form either conjugate or non-conjugate pairs. Non-conjugate pairs occur in a number of famous models, such as CTM and DTM \cite{Lafferty2006,Blei2006a}. The challenge with non-conjugate priors, however, is that the posterior does not belong to the same family as the prior, and often the posterior cannot be obtained in a closed form analytically \cite{Wainwright2007a}. As a result, models with non-conjugate terms often require the introduction of additional variational parameters and gradient based optimizations which significantly slows down training.

{\bf Advances in Variational Inference} In recent years tremendous progress has been made in improving both the speed, quality, and ease of application of variational inference. In 2013, Hoffman et al. introduced stochastic variational inference (SVI): a method that reparameterizes the gradient of the Expected Lower BOund (ELBO) in terms of the natural parameters in order to derive a fast stochastic gradient descent (SGD) algorithm for variational inference \cite{Hoffman2010,Hoffman2013}. The SVI approach is an important contribution because of the tremendous speed-up that results. Further, reparameterizing the ELBO so as to derive natural gradients, which leads to stochastic optimization, is closely related to traditional coordinate ascent variational inference. Natural gradients provide more stable learning, as opposed to gradient methods that are better suited for optimization in Euclidean geometry \cite{amari1998natural}. SVI is limited in that it requires the model's parameters to have an exponential family form, and hence is not directly applicable to non-conjugate models like CTM or DTM.

%Stochastic updates have been shown to speed up convergence of approximate inference algorithms significantly. One reason for this is that full batch gradient updates during the initial iterations inefficiently attempt to infer local variational parameters over the entire dataset given random initializations of the global parameters. Mandt et al. proposed averaging over a fixed number of previous biased stochastic gradients \cite{Mandt2014}, as opposed to following unbiased stochastic gradients as is done in SVI. The bias in the smoothed stochastic gradients follows from the size of fixed window of averaged gradients. While introducing a bias, the averaged gradients have the advantage of reducing the variance.

Recent advances, like Black Box Variational Inference (BBVI), demonstrate a promising new way to speed-up updates to non-conjugate terms \cite{Ranganath2013a}. The approach introduced in BBVI uses stochastic gradient updates, where the noisy stochastic observations are computed using Monte Carlo techniques. BBVI results in a variational inference algorithm that is faster than conventional approaches and eliminates the need to derive inference algorithms for new models.

{\bf Conjugate-Computation Variational Inference} The computational downside of BBVI is that does not take advantage of conjugate terms with closed form updates. Khan and Lin introduce Conjugate-computation Variational Inference (CVI) which cleverly allows inference on models with non-conjugate terms to be computed as a \emph{conjugate computation} \cite{Khan2017}. A conjugate computation is simply the adding of the natural parameters of a prior to the sufficient statistics of the likelihood. In short, the CVI approach allows for fast updates to complex PGMs. Moreover, unlike SVI which takes gradients of the ELBO in the natural-parameter space, CVI uses stochastic mirror descent in the mean-parameter space that eschews Euclidean geometry (i.e. squared loss) in favor of a Bregman divergence defined by the convex-conjugate of the log-partition function. Khan and Lin \cite{Khan2017} demonstrate that this approach leads to inference in a conjugate model, where non-conjugate terms have been replaced by exponential family approximations. Further, CVI lets models be trained with stochastic mini-batches --- in the style of SVI. As a result, even complex models with difficult non-conjugate terms can be trained quickly and efficiently.

The goal of the CVI algorithm is to maximize a lower bound to the marginal likelihood:

\begin{equation*}
\argmax_{\lambdab \in \Lambdab} \mathcal{L}(\lambdab) = \E_q[ \log p(\y, \z) - \log q(\z \mid \lambdab) ]
\end{equation*}

\noindent where $\Lambdab$ is the set of valid variational parameters, $\lambdab$ the variational parameter, and $q(\z \mid \lambdab)$ the variational approximation. Traditionally, the bound is optimized via gradient descent, i.e. $\lambdab_{i+1} \leftarrow \lambdab_ + \rho_i \nabla_{\lambdab} \mathcal{L}(\lambdab_i)$, where $\rho_i$ is the learning rate. An equivalent formulation of this gradient, which highlights the divergence function, is:

\begin{equation*}
\lambdab_{i+1} \leftarrow \argmax_{\lambdab \in \Lambdab} \langle \lambdab, \nabla_{\lambdab} \mathcal{L}(\lambdab_i) \rangle - \frac{1}{2\rho_i} ||\lambdab - \lambdab_i||^2_2
\end{equation*}

CVI assumes distributions are minimal exponential families, meaning there is a one-to-one mapping between $\lambdab$ and the mean parameters $\mub \in \mathcal{M}$. The lower bound is reparameterized in terms of $\mub$ such that $\wt{\mathcal{L}}(\mub) = \mathcal{L}(\lambdab)$, and mirror descent gradient updates for this bound are derived. Additionally, in making the mean-field assumption, which assumes that the parameters are \emph{posteriori} independent, the gradient update can be expressed as a summation over all nodes $k \in 1, \dots, M$:

\begin{equation}
\begin{split}
\label{eq:max_local_obj}
\max_{\mub} \sum_{k=1}^M \Big[ \Big\langle \mub_k, \widehat{\nabla}_{\mub} \widetilde{\mathcal{L}}(\mub_i) \Big\rangle - \frac{1}{\rho_i} \mathbf{B}_{A^*} (\mub_k || \mub_{k,i}) \Big]~,
\end{split}
\end{equation}

\noindent where $i$ refers to the iteration number, and $\mathbf{B}_{A^*}$ is a Bregman divergence --- such as KL divergence, defined over the convex-conjugate of the log-partition $A^*$. The choice of divergence function is to account for the geometry of the parameter space. Khan et al. prove convergence for the general case of Bregman divergences, even in the stochastic gradient setting \cite{Khan2016}. The maximization in \eqref{eq:max_local_obj} only requires optimizing for a single node $k$ and hence can be done either in parallel, or as a doubly stochastic scheme by randomly picking a term in the summation.

One of the primary results proved by Khan and Lin \cite{Khan2017} is that \eqref{eq:max_local_obj} can be implemented as Bayesian inference in a conjugate model. Their method hinges splitting the joint distribution into non-conjugate and conjugate terms (denoted $\tilde{p}_{nc}(\y, \z)$ and $\tilde{p}_{c}(\y, \z)$, respectively) and replacing the difficult non-conjugate term with an exponential family approximation whose natural parameter is $\wt{\lambdab}_i$. Hence, the posterior is approximated with a variational distribution defined by:

\begin{equation*}
q(\z \mid \lambdab_{i+1}) \propto \exp(\phi(\z), \wt{\lambdab}_i) \wt{p_c}(\y, \z)~,
\end{equation*}

\noindent where $\wt{\lambdab}_i$ is the natural parameter of the exponential-family approximation to $\wt{p}_{nc}$, computed as a weighted sum of the gradients of the non-conjugate term. Khan and Lin \cite{Khan2017} show that the exponential-family approximation's parameter $\wt{\lambdab}_i$ and the variational posterior's parameter $\lambdab$ are updated by:

\begin{subequations}
\label{eq:cvi_updates}
\begin{align}
\wt{\lambdab}_{k,t} &= \sum_{a \in \mathbb{N}_k} \E_{q / k,i}[\etab_{a,k}(\z_{a/k}, \y_{a/k})] + \nabla_{\mu_k} \E_{q_i}[\log \wt{p}_{nc}^{\sim a,k}] \label{eq:cvi_updates1} \\
\lambdab_{i+1} &= (1 - \rho_i) \lambdab_i + \rho_i \wt{\lambdab}_i \label{eq:cvi_updates2}
\end{align}
\end{subequations}

\noindent where $\etab_{a,i}(\z_{a/i}, \y_{a/i})$ are simply the natural parameters for the conjugate parts of the model, and $\mathbb{N}_k$ the local neighborhood containing $\z_i$ and its children. By replacing non-conjugate terms with exponential family approximations, CVI allows even complex models to be trained quickly and efficiently.

%% file: 3_dap.tex
\section{Dynamic Author-Persona Topic Model}
\label{model}

The Dynamic Author-Persona (DAP) topic model is designed for corpora with multiple authors writing over time \cite{Giaquinto2018}. Giaquinto et al. introduce the DAP model and demonstrate its ability to identify common narratives shared by patients and caregivers journaling during a serious health crisis on the website CaringBridge. While the model can produce valuable qualitative results from smaller datasets, it struggles to scale to industrial sized problems.

\begin{figure}
\caption{Graphical representation of the Dynamic Author-Persona topic model (DAP). On top, topic distributions for each persona evolve over time by $\alphab_t | \alphab_{t-1} \sim \mathcal{N}(\alphab_{t-1}, \Sigmab)$. The distribution over words for each topic is $\betab \sim Dir(\eta)$. Each author $a \in \{1, \dots, A\}$ is represented by a distribution over personas defined by $\kappab_a \sim Dir(\omega)$. The distribution over topics for each document $\thetab_d \sim \mathcal{N}(\alphab_t \x_{t,d}, \Sigmab_t)$ is dependent on the persona assignment $\x_{t,d} \sim Mult(\kappab_a)$ for that document's author, and the evolving topic distribution $\alphab_t$. Words, denoted $\w$, are assigned to topics according to the multinomial $\z_{d,n} \sim Mult(\sigma(\thetab_{t,d}))$.}
\label{fig:dap}
\centering
\includegraphics[width=1.0\linewidth]{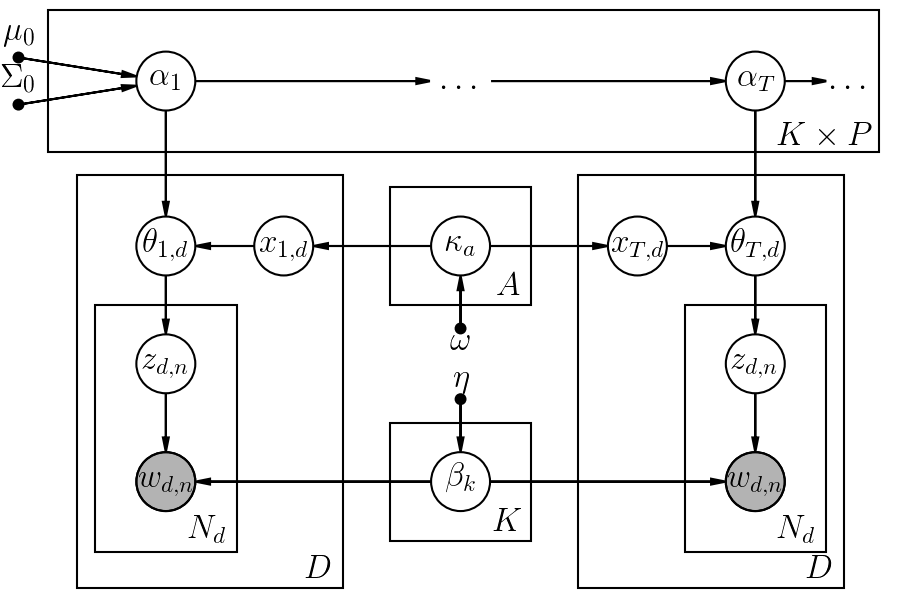}
\end{figure}

To model temporal dependencies between parameters DAP uses a Variational Kalman Filter, similar to \cite{Blei2006a,Wang2008}. The structure of the DAP model, shown in Figure \ref{fig:dap}, is particularly unique due to the parameter $\alphab_{t,p}$, which captures the distribution over topics for each persona $p$ at time point $t$. The structure of the DAP model results in the joint distribution that factorizes as:

\begin{equation}
\begin{split}
\label{eq:dap}
\prod_{k=1}^K & p(\betab_k \mid \eta) \prod_{a=1}^A  p(\kappab_a \mid \omega) \prod_{t=1}^T \prod_{p=1}^P p(\alphab_{t,p} \mid \alphab_{t-1,p}, \Sigmab_{t-1}) \times \\
& \prod_{d=1}^{D_t} p(\x_{t,d} \mid \kappab_{a_d}) p(\thetab_{t,d} \mid \alphab_{t,1:P} \x_{t,d}, \Sigmab_t) \prod_{n=1}^{N_{d_t}} p(\z_{d,n} \mid \sigma(\thetab_{t,d})) ~,
\end{split}
\end{equation}

\noindent where $\sigma(\cdot)$ is a softmax function introduced to obey the constraint that $\z_{d,n}$ lies on the simplex.

The structure and parameterization of the model, however, introduces a number of non-conjugate terms, namely $p(\z_{d,n} \mid \sigma(\thetab_{t,d}))$ and $p(\thetab_{t,d} \mid \alphab_t \x_{t,d}, \Sigmab_t)$. Consequently, estimating the topic assignment $\z$, topic proportions $\thetab$, and persona assignment $\x$ is challenging. The remaining model terms, the variational parameter used in the mean-field variational inference algorithm, and a brief description is given in Table \ref{tbl:dap}.

\begin{table}[ht]
\caption{Notation and parameters used in the DAP model. Variational refers to the corresponding parameter in the mean-field variational inference algorithm.}
\label{tbl:dap}
\centering
\begin{tabular}{llll}
  \toprule
Parameter & Variational & Description \\
  \midrule
  $\w_{t,d}$ &  & Words in document $d_t$ \\
  $\z_n$ & $\phib_n$ & Assigns word $n$ to a topic \\
  $\thetab_{t,d}$ & $\gammab_{t,d}$ & Topic distribution for document $d_t$ \\
  $\mathbf{v}_{t,d}$ & $\hat{\mathbf{v}}_{t,d}$ & Covariance between topics for $d_t$ \\
  $\mu_0$ &  & Prior for mean of $\alphab_0$ \\
  $\Sigma_0$ &  & Prior for covariance of $\alphab_0$ \\
  $\alphab_{t,p} $ & $\hat{\alphab}_{t,p}$ & Persona $p$'s topic distribution  \\
  $\Sigmab_t$ & $\hat{\Sigmab}_t$ & Covariance in topic distributions \\
  $\omega$ &  & Prior for $\kappab_a$ \\
  $\kappab_a$ & $\deltab_a$ & Author $a$'s personas distribution \\
  $\x_{d,t}$ & $\taub_{t,d}$ & Assigns author of $d_t$ to a persona  \\
  $\eta$ &  & Prior parameter for $\beta_k$ \\
  $\betab_k$ & $\lambdab_k$ & $\forall k$ distribution over words  \\
  \bottomrule
\end{tabular}
\end{table}

The DAP model's scalability issues stem from its non-conjugate terms --- for which there are no fast, closed form updates. To derive parameter updates the DAP model's intractable posterior is approximated with a variational posterior under the mean-field assumption. In standard fashion, the Evidence Lower BOund (ELBO) is maximized, which is equivalent to minimizing the KL divergence between the variational in true posteriors. Once the ELBO is specified, updates are derived for each parameter by selecting terms containing that parameter and optimizing. However, due to the non-conjugate terms, fast, closed-form updates are not always possible. In particular the DAP model's E-Step --- which runs multiple times for each document in the corpus --- must use exponentiated gradient descent to learn the persona assignment $\taub$ of an author, and conjugate-gradient descent to learn the mean and variance parameters of the document's topic distribution.

%% file: 4_cvi_updates.tex
\section{DAP Performed Exceedingly Rapidly}
\label{cvi}

The non-conjugate terms in the DAP model compromise the scalability of the model. CVI presents an opportunity to directly address DAP's bottlenecks, while keeping the existing closed-form parameter updates. We refer to DAP trained with the new CVI based inference algorithm as Dynamic Author-Persona Performed Exceedingly Rapidly (DAPPER). Details of the derivation of DAPPER's inference algorithm are shared below. In particular, the algorithm is structured like variational EM, where local variational parameters are updated in the Expectation step on a mini-batch or the entire corpus, and then global model parameters are updated in the Maximization step.

\subsection{E-Step}
In the E-step we update each document $d$'s topic proportions $\thetab_d$, the assignment of each word to a topic $\z_n$, and the assignment of each author to a persona $\x_d$.

\subsubsection{Document Topic Proportions}
Each document is given a hidden parameter $\thetab_d$ representing the proportions of each topic. We begin by identifying the conjugate and non-conjugate terms involving $\thetab_d$. For $\thetab_d$ we have a conjugate term: $\tilde{p}_{c}^{\theta} = \mathcal{N}(\thetab_d \mid \alphab_t \x_d, \Sigma_t)$. Here $\mathcal{N}(\thetab_d \mid \alphab_t \x_d, \Sigma_t)$ is a Gaussian conditioned on a Gaussian because the mean parameter is the dot product between two fixed terms $\alphab_t \x_d$, giving a natural parameter:

\begin{equation}
\label{theta_nat}
    \begin{bmatrix}
        \Sigma_t^{-1}(\alphab_t \x) & -\frac{1}{2}\Sigma_t^{-1} \\
    \end{bmatrix}^\top
\end{equation}

The second term involving $\thetab$ required is the non-conjugate term, which is $\tilde{p}_{nc}^{\theta} = Mult(\z_n \mid \sigma(\thetab_d))$

The variational distribution is defined $q(\theta_{d,k}) = \mathcal{N}(\theta_{d,k} \mid \gamma_{d,k})$ where $\gamma_{d,k} = \{ m_k, v_k\}$ for topic $k$ has sufficient statistics:

\[
ss(\thetab_{d,k}) = \begin{bmatrix} \theta_{d,k} & \theta_{d,k}^2 \\ \end{bmatrix}^\top
\]

Writing the approximate posterior $q(\thetab)$ as a product of the conjugate and non-conjugate parts gives:

\begin{equation*}
\begin{split}
q_{i+1}(\thetab) \propto \Big[ \prod_{k=1}^K \exp( ss(\theta_{d,k}) \wt{\Thetab}_{k,i} ) \Big] \mathcal{N}(\thetab_d \mid \alphab_t \x_d, \Sigma_t)
\end{split}
\end{equation*}

\noindent where $\wt{\Thetab}_{k,i}$ are the natural parameters to the approximated exponential family at iteration $i$.

The difficult non-conjugate term $Mult(\z_n \mid \sigma(\thetab))$ has already been approximated in \cite{Lafferty2006}, specifically:

\begin{equation}
\begin{split}
\label{theta_f}
f &= \E_q \Big[ \log Mult(\z_n \mid \sigma(\thetab)) \Big] \\
&\geq \sum_{k=1}^K m_k \phi_n^{(k)} - \zeta^{-1} \sum_{k=1}^K \exp(m_k + \frac{v_k}{2}) - \log(\zeta) + 1
\end{split}
\end{equation}

\noindent where, again, the parameter $\zeta$ is introduced to preserve a lower bound. Thus, we can now take the gradient of $f = \E_q[ \log Mult(\z_n \mid \sigma(\thetab)) ]$ with respect to the mean parameters.

\begin{equation*}
\nabla_{\mu} f = \begin{bmatrix} \frac{\partial f}{\partial \mu^{(1)}} & \frac{\partial f}{\partial \mu^{(2)}} \\ \end{bmatrix}^\top
\end{equation*}

Since the mean variational distribution's mean parameters are $\mub = \begin{bmatrix} m_k & m_k^2 + v_k \\ \end{bmatrix}^\top$, we can write $m_k = \mu_k^{(1)}$ and $v_k = \mu_k^{(2)} - (\mu_k^{(1)})^2$. By the chain rule the gradient with respect to the mean parameters are $\frac{\partial f}{\partial \mu^{(1)}} = \frac{\partial f}{\partial m} -2 \frac{\partial f}{\partial v}$ and $\frac{\partial f}{\partial \mu^{(2)}} = \frac{\partial f}{\partial v}$. Applying these gradients to the non-conjugate term in \eqref{theta_f} we can then compute the natural parameter of the variational posterior:

\begin{equation}
\begin{split}
\label{theta_grad}
\frac{\partial f}{\partial m} - 2 \frac{\partial f}{\partial v} m &= \phi_n^{(k)} \quad\mathrm{and}\quad
\frac{\partial f}{\partial v_k} = \frac{1}{2 \zeta} \exp(m_k + \frac{v_k}{2}) \\
\implies &\nabla_{\mub}(f) = \begin{bmatrix} \phi_n^{(k)} & -\frac{1}{2 \zeta} \exp(m_k + \frac{v_k}{2}) \\ \end{bmatrix}^\top
\end{split}
\end{equation}

\noindent where $\zeta$ has the same update as before: $\zeta \leftarrow \sum_{k=1}^K \exp(m_k + \frac{v_k}{2})$. Thus by the CVI update rules given in \eqref{eq:cvi_updates}, the natural parameter of the topic proportions is given by a \emph{conjugate computation} adding \eqref{theta_grad} (the sufficient statistics) to \eqref{theta_nat} (the natural parameter of prior):

\begin{equation}
\label{theta_cvi}
\Thetab_{k,i+1} = \rho_i \begin{bmatrix} \sum_{n=1}^{N_d} \phi_n^{(k)} + \Sigma_t^{-1}(\alphab_t \x_d)_k  \\ \frac{-N}{2 \zeta} \exp(m_k + \frac{v_k}{2}) + (-\frac{1}{2} \Sigma_{t, k,k}^{-1}) \\ \end{bmatrix} + (1 - \rho_i) \Thetab_{k,i} \\
\end{equation}

Ideally we want to compute the source parameters to the variational posterior, i.e. $m_k$ and $v_k$ --- which is straight-forward\footnote{These follow from the definitions for converting a Multivariate Gaussian between its source and natural parameters, which can easily be looked up, see for example \cite{Nielsen2009}.} given the natural parameter $\Thetab_{k,i+1}$ computed in \eqref{theta_cvi}. The final updates to the variational posterior's source mean and variances are computed:

\begin{equation*}
m_{k,i+1} = \frac{-\Theta_{k, i+1}^{(1)}}{2 \Theta_{k, i+1}^{(2)}} \quad\mathrm{and}\quad v_{k, i+1} = \frac{-1}{2 \Theta_{k, i+1}^{(2)}}
\end{equation*}

\subsubsection{Topic Assignment}
Each word is assigned to a topic through a hidden parameter $\z_n$. For $\z_n$ the corresponding conjugate term is $\tilde{p}_{c}^{\z} = Mult(\w_n \mid \betab_{z_n})$, and non-conjugate term is $\tilde{p}_{nc}^{\z} = Mult(\z_n \mid \sigma(\thetab))$.

Define the variational distribution $q(\z_n) = Mult(\z_n \mid \phi_n)$ for word $n$. Writing the approximate posterior $q(\z_n)$ as a product of the conjugate and non-conjugate parts:

\begin{equation}
\begin{split}
\label{q_z}
q_{i+1}(\z_n) \propto \Big[ \prod_{n=1}^{N_d} \exp( ss(\z_n) \wt{\Phib}_{i} ) \Big] Mult(\w_n \mid \betab_{z_n})
\end{split}
\end{equation}

\noindent where $\wt{\Phib}_{i}$ are the natural parameters to the approximated exponential family at iteration $i$, computed by:

\begin{equation*}
\begin{split}
\wt{\Phib}_{i} &= \begin{bmatrix} \wt{\Phi}_{1,i} & \dots & \wt{\Phi}_{K,i} \\ \end{bmatrix}^\top \\
&= \rho_i \nabla_{\phi} \E_{q_i} \Big[ \log Mult(\z_n \mid \sigma(\thetab)) \Big] |_{\phi = \phi_i} + (1 - \rho_i) \wt{\Phib}_{i-1}
\end{split}
\end{equation*}

Using the previously computed approximation to the challenging term $\log Mult(\z_n \mid \sigma(\thetab))$ in \eqref{theta_grad}, we now differentiate it with respect to the mean parameter of $q(\z_n)$, i.e. $\phi$. This gives:

\begin{equation}
\begin{split}
\nabla_{\mub} \E_{q_i} \Big[ \log Mult(\z_n \mid \sigma(\thetab)) \Big] &= \begin{bmatrix}
        m_1 & \dots & \dots m_K \\
    \end{bmatrix}^\top = \m \\
\end{split}
\end{equation}

\noindent where $m_k$ is the mean of $q(\theta_{d,k}) = \mathcal{N}(\theta_{d,k} \mid m_k, v_k)$, for $k \in 1, \dots, K$. To update the natural parameter to the approximated exponential family in \eqref{q_z}, we use the equations:

\begin{equation}
\begin{split}
\wt{\Phib}_{i} &= \rho_i \Big( \nabla_{\mub} \E_{q_i}[\log Mult(\z_n \mid \sigma(\thetab))] \Big) + (1-\rho_i)\wt{\Phib}_{i-1} \\
&=\rho_i \m + (1-\rho_i)\wt{\Phib}_{i-1}
\end{split}
\end{equation}

Since CVI transforms non-conjugate computations into conjugate computations the update to the variational parameter $\phib_{n}$ is similar to the LDA case. Specifically, we compute the source parameter of our variational posterior $\phib_n$ by:

\begin{equation}
\begin{split}
\phi_{n,k,i+1} \propto \exp( \wt{\Phi}_{k,i} + \E_q[\log \beta_{k,w_n}] )
\end{split}
\end{equation}

\noindent where, as usual, the Dirichlet expectation $\E_q[\log \beta_{k,v}]$, is computed by $\Psi(\lambda_{k,v}) - \Psi(\sum_{j=1}^V \lambda_{k,j})$. The first term $\wt{\Phi}_{k,i}$ is the document's topic distribution computed in the previous section, hence (as expected) the CVI update for $\phib$ results in a simple closed form update with the same form as in the original DAP model.

\subsubsection{Persona Assignment}
For each document the author is assigned to a persona through a hidden parameter $\x_d$. For $\x_d$ the conjugate term is $\tilde{p}_{c}^{\x} = Mult(\x_a \mid \kappa_{d_a})$, and its non-conjugate term is $\tilde{p}_{nc}^{\x} = \mathcal{N}(\thetab \mid \alpha_t \x_d, \Sigma_t)$, where the coupling in the mean $\alpha_t \x_d$ made closed form updates in the DAP model impossible.

Define the variational distribution $q(\x_d) = Mult(\x_d \mid \tau_d)$. Writing the approximate posterior $q(\x_d)$ as a product of the conjugate and non-conjugate parts gives:

\begin{equation}
\begin{split}
\label{q_x}
q_{i+1}(\x_d) \propto \Big[ \prod_{p=1}^{P} \exp( ss(\x_d) \wt{\tau}_{d,p,i} ) \Big] Mult(\x_d \mid \kappab_{d_a})
\end{split}
\end{equation}

\noindent where $\wt{\taub}_{d,i}$ are the natural parameters to the approximated exponential family at iteration $i$.

We compute the variational posterior by first taking the gradient of the non-conjugate terms, where the non-conjugate term $f = \E_{q_i}[\mathcal{N}(\thetab \mid \alpha_t \x_d, \Sigma_t)]$ evaluates to:

\begin{equation*}
\begin{split}
f &= \frac{-1}{2} \Big( (\gamma_{t,d} - \hat{\alpha}_{t} \tau_{t,d})^\top \Sigma_t^{-1} (\gamma_{t,d} - \hat{\alpha}_{t} \tau_{t,d}) +\\
& \sum_{p=1}^P \Tr \Big[ \Sigma_t^{-1} \diag \Big( \tau_{t,d,p} (\hat{\alpha}_{t,p} \hat{\alpha}_{t,p}^\top + \hat{\Sigma}_{t}) \Big) \Big] \Big) + const\\
\end{split}
\end{equation*}

\noindent Taking the gradient with respect to the mean parameter $\taub$ gives:

\begin{equation*}
\begin{split}
\nabla_{\tau} f &= \hat{\alpha}_{t, p} \Sigma_t^{-1} (\gamma_{t,d} - \hat{\alpha}_{t, p} \tau_{t,d, p}) - \frac{1}{2} \Tr(\Sigma_t^{-1} \diag(\hat{\alpha}_{t,p}^2 + \hat{\Sigma}_{t})) \\
\end{split}
\end{equation*}

Next we use the CVI updates rule from \eqref{eq:cvi_updates} to compute the natural parameter of our variational posterior $q(\x_d)$ by:

\begin{equation*}
\begin{split}
\wt{\taub}_{d,i} &= \rho_i \Big( \E_q[\log \kappab_{d_a}] + \nabla_{\tau} f \Big) + (1-\rho_i)\wt{\taub}_{d,i-1} \\
\end{split}
\end{equation*}

\noindent where $\E_q[\log \kappab_{d_a}]$ is the expected natural parameters from the conjugate term, and is equivalent to a Dirichlet expectation: $\Psi(\delta_{a,p}) - \Psi(\sum_{j=1}^P \delta_{a,j})$. In order to map the natural parameter $\wt{\taub}_{d,i}$ back to the source parameter of the variational posterior $q(\x_d)$, we use

\begin{equation*}
\begin{split}
\taub_{d,i} &= \begin{bmatrix}
                \frac{\exp( \wt{\taub}_{d,1,i} )}{\sum_{p=1}^P \wt{\taub}_{d,p,i}} & \dots & \frac{\exp( \wt{\taub}_{d,P,i} )}{\sum_{p=1}^P \wt{\taub}_{d,p,i}} \\
            \end{bmatrix}^\top \\
\end{split}
\end{equation*}

\subsection{M-Step}
In the M-step we use sufficient statistics collected from computing document-level variational parameters computed during the E-step to update the global parameters $\betab, \kappab$, and $\alphab$. Because DAPPER makes use of stochastic mini-batches, we use the learning rate defined for SVI and recommended in CVI: $\rho_i = (i + \tau)^{-\kappa}$ where $\tau \geq 0$ is the delay and $\kappa \in (0.5, 1.0]$ is the forgetting rate \cite{Hoffman2010,Hoffman2013,Khan2017}.

\subsubsection{Topic's Distribution over Words}
The $\betab$ term is already conjugate, and hence the variational distribution for the topics, $q_{i+1}(\betab) = \prod_{k=1}^K Dir(\betab_k \mid \lambdab_{k, i+1})$, already has a closed form update: $\lambdab_{k,i+1} = (1 - \rho_i)\lambdab_{k, i} + \rho_i (\eta + \sum_{d=1}^D \sum_{n=1}^{N_d} \phi_{d,n,k} w_{d,n})$.

\subsubsection{Author's Distribution over Personas}

Since the $\kappab$ terms are already conjugate and have closed form solutions, it follows that the update to variational posterior, $q_{i+1}(\kappab) = \prod_{d=1}^D Dir(\kappab_{d_a} \mid \deltab_{d_a, i+1})$, has a simple closed form solution using the convex combination: $\deltab_{d_a,i+1} = (1 - \rho_i) \deltab_{d_a, i} + \rho_i (\omega + \sum_{d=1}^D \taub_d)$.

\subsubsection{Persona's Distribution over Topics}
The $\alphab_t$ term is conjugate to all other factors, and is global. The variational distribution for the distribution over topics for each persona $\alphab_{t, 1:P}$ is $q_{i+1}(\alphab_{t,p}) = \prod_{t=1}^T \mathcal{N}(\alphab_{t,p} \mid \hat{\alphab}_{t-1,p, i+1}, \Sigma_t) \prod_{d=1}^D \mathcal{N}(\thetab_d \mid m_d, v_d)$. As shown in the original derivation of the DAP model, a closed form update can be found for $\hat{\alphab}_{t,p}$:

\begin{equation}
\label{alpha_hat_gradient}
\hat{\alphab}_{t,p}^{new} = \frac{\hat{\alphab}_{t-1,p} + \sum_{d=1}^{D_t} \gammab_{t,d}\tau_{t,d,p} - \sum_{d=1}^{D_t} \tau_{t,d,p}}{1 + \sum_{d=1}^{D_t}  \tau_{t,d,p}^2}
\end{equation}

Thus, for mini-batch training we update $\hat{\alphab}_{t,p, i+1}$, by first computing $\hat{\alphab}_{t,p}^{*}$ from a mini-batch of documents. Then $\hat{\alphab}_{t,p, i+1}$ is updated via a convex combination: $\hat{\alphab}_{t,p, i+1} = (1 - \rho_i) \hat{\alphab}_{t,p, i} + \rho_i \hat{\alphab}_{t,p}^{*}$. Alternatively, to encourage personas to be distinct the update \eqref{alpha_hat_gradient} is replaceable by the Regularized Variational Inference (RVI) update for $\alphab_{t,p}$ given in the DAP model. CVI compliments RVI because closed form updates, such as the $\hat{\alphab}_{t,p}$ update found by the regularized DAP model, are preserved. After computing $\hat{\alphab}_{t,p, i+1}$, we proceed as usual and apply the forward and backward equations of the variational Kalman Filter to smooth over time time steps.

\subsection{Connect Between CVI and Expectation Propagation} While DAPPER's inference algorithm is based on CVI --- a recent advance in approximate inference, CVI itself has theoretical connections to the well known expectation propagation (EP) algorithm\cite{Minka2001,Minka2001a}. The EP algorithm, which is an extension of Assumed Density Filtering, infers the approximate posterior $q$ using localized inferences. With posterior $p$, hidden parameters $\z$, and observations $\y$, we assume $p$ can be written as a product of terms: $p(\z \mid \y) \propto \prod_{i=0}^N f_i(\z)$, where $f_0(\z) = p(\z)$ expresses the prior density and $f_i(\z) = p(\y \mid \z)$ the likelihood. The EP algorithm then approximates the posterior, choosing an approximating family with density $q(\z) \propto \prod_{i=1}^N q_i(\z)$ and iteratively incorporating $q_i(\z)$ into $q(\z)$. First, EP computes the cavity distribution --- that is, deleting $q_i(\z)$ from $q(\z)$, by $q_{-i}(\z) \propto q(\z) / q_i(\z)$. Second, a true Bayesian update incorporates $f_i(\z)$:

\begin{equation*}
\hat{p}(\z) = Z_i^{-1} q_{-i}(\z) f_i(\z)~, \quad Z_i = \E_{\z \sim q_{-i}}[f_i(\z)]
\end{equation*}

\noindent where $\hat{p}(\z)$ is a tilted exponential family. The exact posterior is approximated, for exponential families minimizing KL divergence between the posteriors,

\begin{equation}
\label{eq:kl}
q^{new}(\z) = \argmin KL(\hat{p}(\z) || q(\z))~,
\end{equation}

\noindent corresponds to matching the moments of $p$ and $q$. Finally, update $q_i^{new}(\z)$ by $q_i^{new}(\z) \propto q(\z) q_{-i}(\z)$. Note, that the update to $q_i(\z)$ is the local minimization and can be formulated as:

\begin{equation*}
q_{i}^{new}(\z) = \argmin KL(f_i(\z) q_{-i}(\z) || q_i(\z) q_{-i}(\z)).
\end{equation*}

From here two connections to CVI appear. First, EP also computes an exponential family approximation in its approximation of $\hat{p}(\z)$, the tilted distribution induced by $f_i(\z)$ \cite{seeger2003bayesian}. While EP does this computation using moment matching, moment matching corresponds to minimizing the Kullback-Leibler divergence from the tilted distribution to the new approximated marginal distribution \cite{Gelman2014}, and moments can be computed as derivatives of the log normalizer, hence:

\begin{equation*}
\begin{split}
\mu^{new} = \E_{\hat{P}}[\phi(\z)] &= \nabla_{q_{-i}} \log Z_i + \mu_{-i} \\
&= \nabla_{q_{-i}} \log \E_{\z \sim q_{-i}}[f_i(\z)] + \eta_{-i}
\end{split}
\end{equation*}

For exponential families $\nabla \log Z = E[\phi(\y)]$, and therefore this moment matching can be viewed as similar to conjugate computations, here we add expectations of sufficient statistics of $q$ to the corresponding expectations of $\z$ in $q_{-i}(\z) f_i(\z)$. This shows that EP's creation of the tilted distribution moment parameter is analogous to CVI's computation of the natural parameter to the exponential family approximation, i.e. \eqref{eq:cvi_updates1}.

The second connection to CVI lies in the ``damping'' technique used to improve the convergence of EP. Damping replaces the generic update $\lambda_i \rightarrow \lambda_{i+1}$ by a convex combination, which reduces the step size so that only a partial update is applied. CVI also updates the natural parameters with a convex combination: $\lambda_{t+1} = (1-\beta)\lambda_t + \beta \wt{\lambda}_t$  in CVI is essentially just ``damped'' updates in EP \cite{Gelman2014}. This form of updating is analogous to minimizing an alpha divergence (which includes directed KL as a special case).

Despite a number of similarities, EP and CVI differ critically in convergence guarantees. Khan et al. show that CVI converges under fairly mild assumptions, namely that $q$ is a minimal exponential family and the model's conditional distribution can be split into conjugate and non-conjugate terms. EP, on the other hand, is not guaranteed to converge. EP minimizes KL divergence for each local observation, but does not directly minimize $KL(p \mid \mid q)$.

%% file: 5_experiments.tex
\section{Experiments}
\label{experiments}

To evaluate the performance of DAPPER we perform a quantitative comparison with similar topic models (LDA, DTM, CDTM, and DAP), and a qualitative demonstration of DAPPER's scalability and output on the CB\footnote{CB data were acquired with the permission and collaboration of CB leadership in accordance with CB's Privacy Policy \& Terms of Use Agreement. Because of their highly sensitive content the CB dataset has been anonymized, but deidentification techniques are imperfect\cite{narayanan2010myths} and hence we cannot publicly release the CB dataset. Those interested in the dataset are encouraged to contact the investigators. All code for training the DAPPER model and running our experiments on the SM dataset, however, are available at \url{https://github.com/robert-giaquinto/dapper}.} and Signal Media One-Million News Article\footnote{\url{http://research.signalmedia.co/newsir16/signal-dataset.html}} (SM) corpora \cite{Signal1M2016}. For the quantitative comparison per-word log-likelihoods ($PWLL$) are computed on test data, where $PWLL = \frac{\sum_{d=1}^D \log p(\w_d)}{\sum_{d=1}^D N_d}$. While $PWLL$s do not correlate with a model's ability to discover coherent topics \cite{Chang2009}, they do offer a fair comparison of how well each model optimizes its objective function. Additionally, the speed and efficiency of DAPPER relative to its predecessor are measured by showing model performance as a function of training time. The qualitative comparison demonstrates the rich and compelling ``health journeys'' discovered by DAPPER on the CB corpus.

\subsection{Datasets}
% names and preprocessing

Both the CB and SM corpora are pre-processed by removing common stopwords and reducing words to their lemma forms. Document timestamps are converted into a continuous, relative measure; for CB we use the number of weeks since author's first post (only looking at the first year of each authors journals), and for SM we use the day within span of the corpus (1-30 September, 2015).

% cb + full cb
{\bf CaringBridge.} Established in 1997, CaringBridge is a 501(c)(3) non-profit organization that connects people and reduces the feelings of isolation that are often associated with a patient's health journey. DAPPER and its predecessor are designed with the CaringBridge corpus in mind. We demonstrate scalability and quality of DAPPER's output on the full CB corpus. The CaringBridge corpus consists of 9,010,623 journals written by 200,388 authors (with a total of 937,503,945 words) between 2006 and 2016 on the CaringBridge website. On average, authors write 100 words per journal and 45 journal posts in the first year.

% qual cb
For a qualitative evaluation, 22,552 randomly selected CB journals are set aside as a test set to evaluate the model and track convergence, leaving 8,988,071 journals in the training set. Our goal in training the DAPPER model on this dataset is to show that the model can find compelling qualitative results even on massive, complex datasets.

% quant CB
A quantitative evaluation on a subset of CB journals is drawn from 2,000 randomly selected authors, leaving a total of 114,532 journals. We refer to this corpus as CB-subset. From here 90\% of the journals ($N=103,018$) are divided into the training set, and the remaining 10\% of each author's journals ($N=11,728$) make up the test set. Training and test sets contain the same authors because personas distributions are learned for each author during training. These authors journal an average of 57 times during the first year, with a mean of 5 days between journal posts.

% sm data
{\bf Signal Media Blogs.} From the SM dataset we only consider articles written by bloggers who wrote fewer than one blog post per day during the corpus' one month span. Subsetting the data in this way is done to exclude major news organizations and instead focus on bloggers who typically write about a central theme. We refer to the subset as SM-blogs. After pre-processing, the SM-blogs corpus consists of 97,839 documents for training (15,848 blogs, 19,278,689 total words), and 10,887 documents for testing (same authors, 2,165,634 words).

\subsection{Hyperparameters}
\label{sec:hyperparameters}
To ensure a fair comparison we fix hyperparameters appearing in each of the models, such as number of topics and convergence criteria. Relative differences between model performances don't vary significantly depending on the number of topics chosen, and hence we only report results for models with 25 topics on the CB-subset and 50 topics on the SM-blogs. DAP and DAPPER seek 15 and 25 latent personas for the CB-subset and SM-blogs corpora, respectively, and fix their regularization of personas to $\rho=0.2$. Because DAPPER can be trained on stochastic mini-batches we report results for various mini-batch sizes and full-batch training.

%% file: 6_results.tex
\section{Results}
\label{results}

\subsection{Model Performance Comparison}
To compare the performance of DAPPER, we train and test DAPPER along with four similar topic models (LDA, DTM, CDTM, and DAP) on the quantitative corpora (see Section \ref{experiments}). Each model is trained for a maximum of 24 hours on a single Haswell E5-2680v3 processor or until training performance converged --- although the DAP model is the only model not to converge within 24 hours. Performance of each model is shown in Table \ref{tbl:overall_results}. The DAPPER model shows significant performance improvements over all competing models due to its faster method for handling non-conjugate terms. Performance for DAPPER is shown for a mini-batch size of 512, which consistently achieved the best training set performance after 24 hours, and DAPPER trained with full batch gradients updates, which achieved the best overall test performance after 24 hours.

\begin{table}
\caption{Overall comparison of models after a maximum of 24 hours of training on CB-subset and SM-blogs corpora. Per-word Log-Likelihoods are reported for the test corpus.}
\label{tbl:overall_results}
\centering
\begin{tabular}{lrr}
\toprule
Model & CB-subset & SM-blogs \\
\midrule
\textbf{DAPPER} (full batch)  & -6.73 & -5.76 \\
DAPPER (batch size = 512)     & -8.19 & -6.31 \\
DAP                           & -8.84 & -7.50 \\
CDTM                          & -8.81 & -8.24 \\
DTM                           & -9.59 & -7.93 \\
LDA                           & -9.23 & -7.79 \\
\bottomrule
\end{tabular}
\end{table}

\begin{table}
\caption{Hours of training for DAPPER to outperform DAP's best performance ($PWLL = -8.65$) on the CB-subset test set. }
\label{tbl:speedup}
\centering
\begin{tabular}{lrr}
\toprule
  Batch Size & Hours to Exceed $PWLL = -8.65$ & Speedup \\
\midrule
  DAP & 40.43 & Baseline \\
  256 & 37.32 & 1.1x \\
  512 & 2.21 & 18.3x \\
  1024 & 1.97 & 20.5x \\
  2048 & 2.13 & 19.0x \\
  4096 & 4.11 & 9.8x \\
  Full Batch & 7.18 & 5.6x \\
\bottomrule
\end{tabular}
\end{table}

Table \ref{tbl:overall_results} highlights three important results: first, the DAP topic model which is designed for the multi-author, temporal structure of the CB dataset achieves competitive performance but clearly suffers by not converging within 24 hours. Second, the DAPPER model benefits from faster training and achieves state-of-the-art performance. Third, smaller mini-batches like 512 result in good training performance that converges quickly but the model does not generalize as well as DAPPER trained with full batch gradients.

In Table \ref{tbl:model_settings} we show the performance of the DAPPER model on the SM-blogs corpus with varying hyperparameter settings. Specifically, we train models with [100, 75, 50, 25] topics, [50, 25, 15] latent personas, and mini-batch sizes of either [256, 512, 1024, 2048] or full gradient training. Full batch training results in the highest per-word log-likelihoods on the test set. Varying the number of personas and the number of topics has a noticable impact on performance (smaller models tend to do slightly better), however batch size is the most significant factor in achieving optimal performance. Smaller models (in terms of number of personas and topics) tend to do well on the 97,839 document SM-blogs corpus, however the best models used full batch training with 50 topics and either 15 or 25 personas.

\begin{table*}
\caption{Comparison of DAPPER's performance on the SM-blogs test corpus for varying number of topics, personas, and batch sizes. In general, we find the full batch training ultimately leads to higher per-word log-likelihoods. For additional hyperparameters, personas has the smallest impact on performance, and the number of topics has noticable impact. The best three models, in terms of highest test set PWLL, are highlighted in bold.}
\label{tbl:model_settings}
\centering
\begingroup\small
\begin{tabular}{ll|rrrrr}
\toprule
Number of Topics & Personas & Batch Size:  256 & 512 & 1024 & 2048 & Full Batch \\
\midrule
25 & 15 & -6.46 & -6.26 & -6.17 & -6.16 & -5.65 \\
25 & 25 & -6.47 & -6.26 & -6.21 & -6.23 & -5.68 \\
25 & 50 & -6.55 & -6.35 & -6.34 & -6.35 & -5.71 \\
50 & 15 & -6.47 & -6.30 & -6.11 & -6.16 & \textbf{-4.97} \\
50 & 25 & -6.50 & -6.31 & -6.30 & -6.39 & \textbf{-5.07} \\
50 & 50 & -6.61 & -6.38 & -6.38 & -6.52 & -5.47 \\
75 & 15 & -6.87 & -6.59 & -6.46 & -6.53 & \textbf{-5.08} \\
75 & 25 & -6.85 & -6.61 & -6.55 & -6.61 & -5.42 \\
75 & 50 & -6.96 & -6.80 & -6.79 & -6.95 & -6.00 \\
100 & 15 & -7.14 & -6.93 & -6.90 & -6.98 & -5.67 \\
100 & 25 & -7.07 & -6.90 & -6.91 & -7.02 & -5.99 \\
100 & 50 & -7.33 & -7.17 & -7.16 & -7.31 & -6.72 \\
\bottomrule
\end{tabular}
\endgroup
\end{table*}

\subsection{Speed and Efficiency}
\begin{figure*}
\caption{Per-word Log-Likelihood performance on the CB training and test sets (larger is better). Each point represents the performance evaluated at the end of an epoch. Each model was trained for a maximum of 48 hours. DAPPER, which incorporates stochastic CVI updates, achieves better likelihoods and converges faster than the DAP model trained with variational EM. Performance of the DAPPER model varies by mini-batch size.}
\label{fig:performance}
\centering
\includegraphics[width=0.9\textwidth]{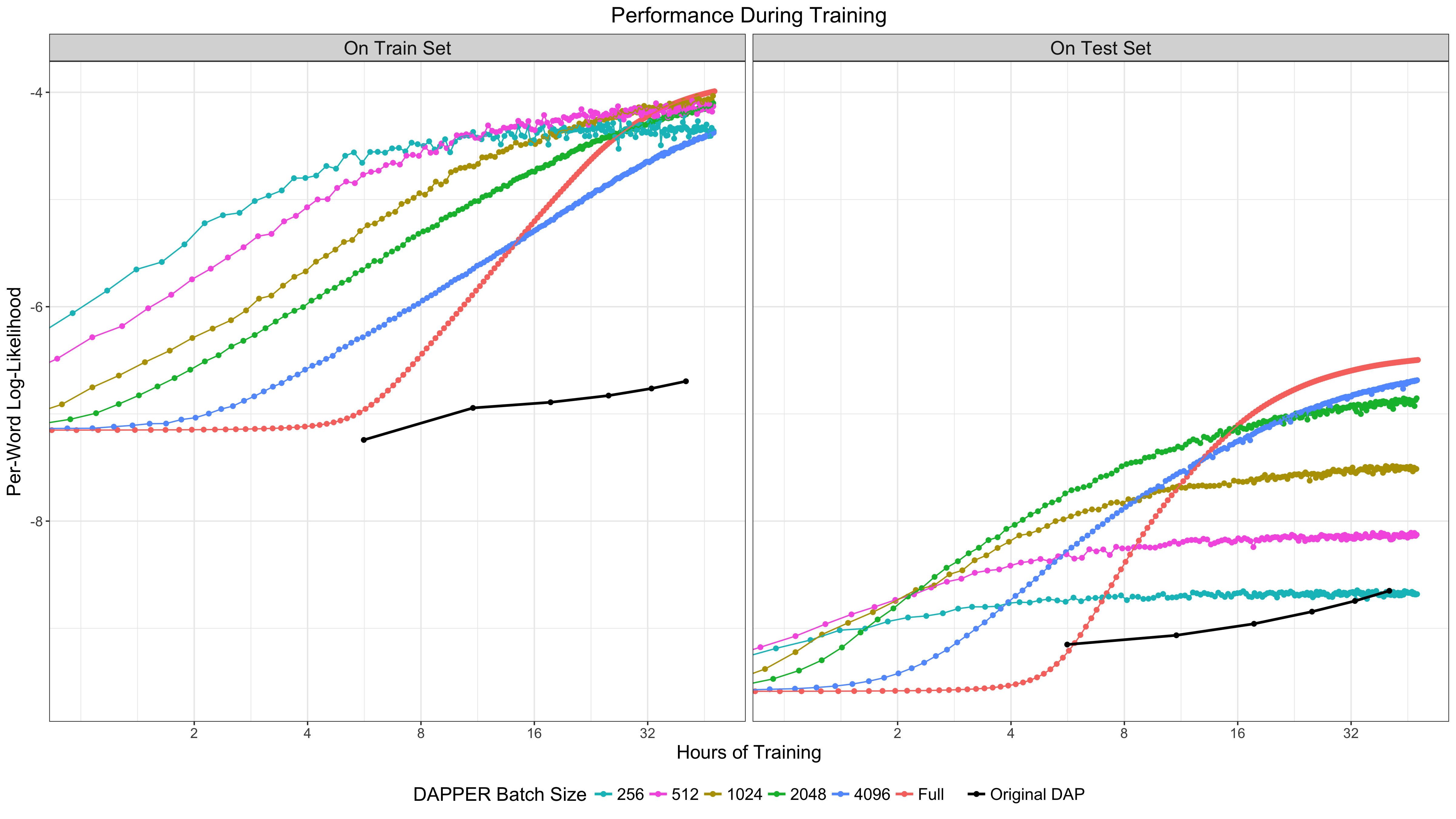}
\end{figure*}

Test set performance of the DAPPER model varies significantly depending on the batch size. Shown in Figure \ref{fig:performance} is the performance of the DAP and DAPPER models trained on the CB-subset corpus and evaluated on the training and test sets after each epoch (one full pass over the training corpus). Each epoch of the DAP model takes an average of 6.7 hours to complete, whereas the DAPPER takes roughly 0.2 hours. The right plot (training set performance) shows that all DAPPER batch sizes begin to converge to a similar value.
The training results (left plot in in Figure \ref{fig:performance}) show that all batch sizes converge to a similar value. On the test set (right plot in in Figure \ref{fig:performance}), however, larger batch sizes show better generalization.

Smaller batch sizes improve quickly at first but ultimately converge to lower PWLLs. The poor performance of small batch sizes may be due to the mini-batches being too noisy. Conversely, the larger batch sizes achieve the best performances, but improve slowly \emph{at first}. We summarize this phenomenon in Table \ref{tbl:speedup}, which reports how quickly DAPPER overtakes the optimal test set performance achieved by DAP. For example, a batch size of 256 converges almost immediately and takes many hours to eventually surpass DAP's best test set result. Whereas a batch size of 1024 improves steadily, and surpasses DAP's best PWLL in a fraction of the time. Despite the implication that the high variance of smaller batch sizes limits performance, we saw no benefit to gradient smoothing techniques, such as those proposed in \cite{Mandt2014}.

\subsection{Scalability and Qualitative Results}
\begin{figure*}
\caption{Selected personas learned by the DAPPER model on the full CaringBridge collection of journals. Each plot shows a different persona, and the three topics most strongly associated with that persona. For clarity, topic labels are hand-defined based on the top words in the topic and journals most associated with that topic (see Table \ref{tbl:words}). Personas show a variety of health journeys. An appeal to a higher power and prayer are common in many journals, and appear in personas 0, 29, and 42. Similarly, a deep reflection on life and death, possibly with respect to one's child appear in 0 and 29. Persona 22 captures a common experience of caring for an aging parent, beginning with intensive care and possibly ending with a hospice or nursing home. Persona 26 shows alternating periods of medical tests and intensive care with times of celebration. Personas 42 and 48 are both associated with cancer, but display very different narratives. Persona 48 includes the pair of topics ``Insurance'' and ``URL Donation'' which often appear together, indicating an author struggling with insurance and medical bills and seeking financial support from friends and family.}
\label{fig:personas6}
\centering
\includegraphics[width=0.8\textwidth]{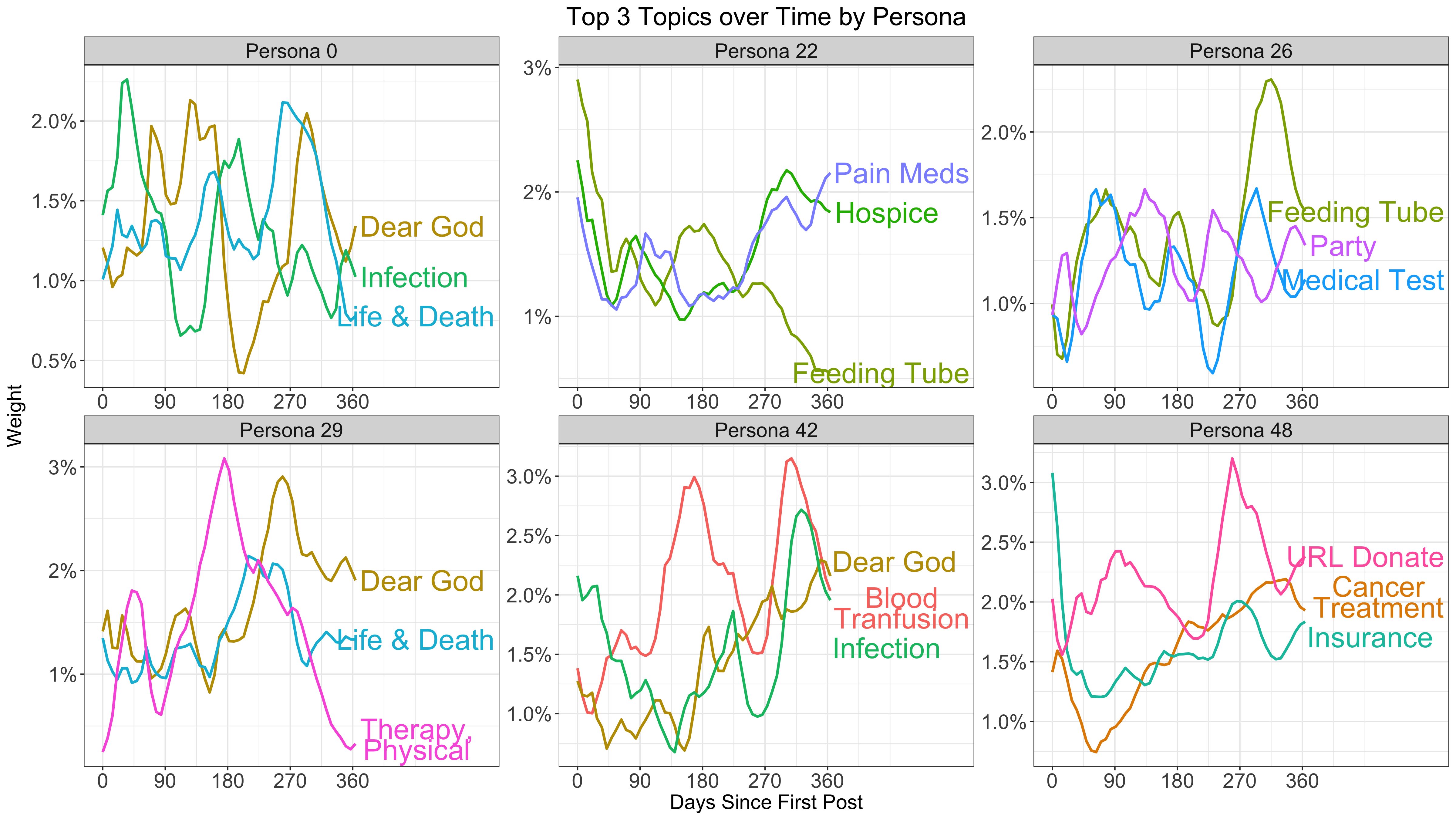}
\end{figure*}

\begin{table*}
\caption{Top eight words associated with the most prevalent topics found by the DAPPER model trained on the full CaringBridge dataset. Topic labels are selected manually in order to aid reference with Figure \ref{fig:personas6}. The words \_dollars\_, \_name\_, and \_URL\_ refer to the result of text pre-processing steps for capturing common patterns like the dollar amounts, anonymized names, and website URLs, respectively.}
\label{tbl:words}
\centering
\begingroup\small
\begin{tabular}{lllllllll}
  \toprule
Infection & Life \& Death & Dear God & Pain Meds & Friend, Memories & Feeding Tube & Party & Medical Test \\
  \midrule
infection & life & god & cause & beautiful & tube & school & dr \\
   fluid &  live &  lord &  pain &  friend &  feed &  birthday &  test \\
   lung &  child &  praise &  medication &  celebrate &  breathe &  fun &  scan \\
   remove &  world &  peace &  brain &  \_name\_ &  weight &  \_name\_ &  result \\
   procedure &  others &  pray &  dose &  card &  oxygen &  party &  drug \\
   chest &  moment &  trust &  increase &  memory &  gain &  aunt &  mri \\
   pressure &  fear &  father &  level &  flower &  rate &  game &  ct \\
   antibiotic &  choose &  joy &  steroid &  dance &  ventilator &  kid &  liver \\
   \bottomrule
 & & & & & & & & \\
  \toprule
Therapy, Physical & Blood Tranfusion & Child & Hospice & Cancer Treatment & URL Donate & ICU & Insurance \\
  \midrule
therapy & blood & play & mom & cancer & \_url\_ & icu & provide \\
   physical &  count &  daddy &  dad &  radiation &  \_dollars\_ &  brain &  medical \\
   leg &  cell &  mommy &  visit &  tumor &  donate &  monitor &  information \\
   therapist &  low &  girl &  visitor &  oncologist &  money &  stable &  disease \\
   arm &  bone &  boy &  hospice &  surgeon &  benefit &  wound &  insurance \\
   rehab &  transplant &  \_name\_ &  nursing &  chemotherapy &  en &  neck &  condition \\
   foot &  white &  little &  facility &  breast & donation &  doctor &  regard \\
   pt &  marrow &  cute &  phone &  biopsy &  ha &  unit &  decision \\
   \bottomrule
\end{tabular}
\endgroup
\end{table*}

To demonstrate the scalability of the DAPPER model, we train DAPPER on the full CB corpus. Figure \ref{fig:personas6} presents selected personas discovered by a DAPPER model with 100 topics and 50 personas. The model is trained using a 24 processor machine for 94 hours, using a regularization of $\rho = 0.15$ and a batch size of 4096. DAPPER's efficient inference algorithm scales to massive datasets. Additionally, with stochastic updates only a constant amount of memory is required. The personas shown in Figure \ref{fig:personas6} highlight a variety of health journeys experienced by CB authors. In Table \ref{tbl:words} we list the most likely words associated with each topic as well as the hand-defined labels assigned to each topic.

Scaling DAPPER to the full CB corpus makes it possible to build larger, richer models --- which in turn can discover a broader range of narratives. Compared to results found by DAP in \cite{Giaquinto2018}, DAPPER's scalability leads to the discovery of many new topics and personas. Many of the new topics discovered by DAPPER are unrelated health conditions. For example, ``Friend, Memories,'' and ``Life and Death'' highlight how authors blend health and life updates in their journaling. This makes sense, a patient's condition is often known by readers or has been previously publicized on the author's homepage, and thus the focus of journals is instead on the patient's current health state. Moreover, health updates tend to focus on procedures (like medical tests and tools), or more general health descriptions like side-effects, infection, pain, or specific body-parts.

\begin{figure*}
\caption{Selected personas learned by the DAPPER model on the SM-blogs corpus. Each plot shows a different persona, and the three topics most strongly associated with that persona. For clarity, topic labels are hand-defined based on the top words in the topic and blog posts most associated with that topic (see Table \ref{tbl:words2}). Results produce by DAPPER trained with full batch gradients, 50 topics, and 15 personas, which was found to perform best during compared to other hyperparameter settings (see Table \ref{tbl:model_settings} for results on additional settings).}
\label{fig:sm_personas}
\centering
\includegraphics[width=0.8\textwidth]{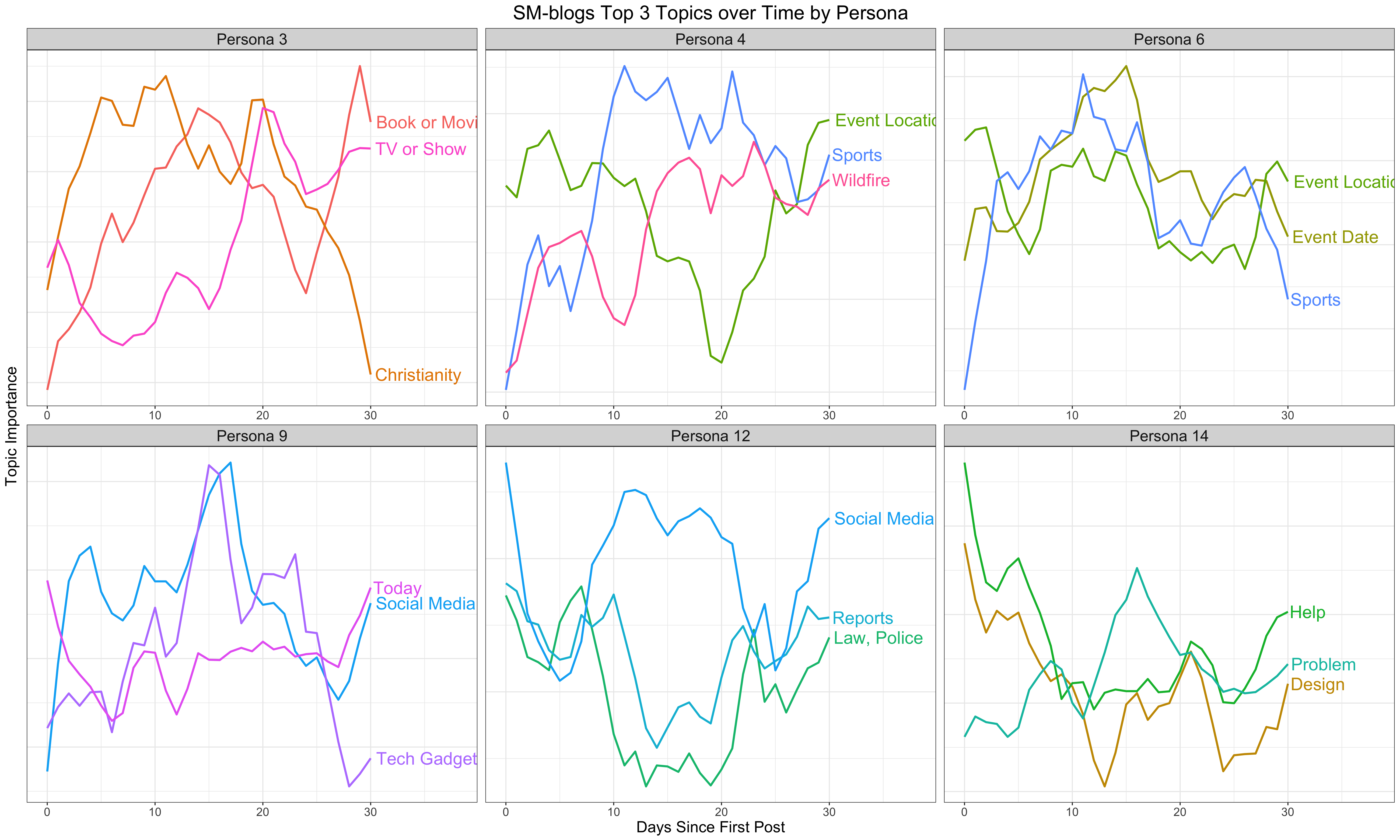}
\end{figure*}

Finally, in Figure \ref{fig:sm_personas} we share qualitative results from DAPPER on the SM-blogs when trained with full batch gradients, 50 topics, and 15 personas --- which was found to perform best during compared to other hyperparameter settings. Figure \ref{fig:sm_personas} shows the top three weighted topics for selected personas, highlighting how DAPPER discovers groups of authors whose writing blends unique topics over time. For instance, authors in Persona 12 tend to talk predominantly about the law and police in additional to reports and public records. Like a number of other personas, Persona 12 often references social media, which is associated with discussing something the author discovered through social media or the author encouraging readers to share and comment on their blog.

\begin{table*}
\caption{Top words associated with the most prevalent topics found by the DAPPER model trained on the full SM-blogs corpus. Topic labels are selected manually in order to aid reference with Figure \ref{fig:sm_personas}.}
\label{tbl:words2}
\centering
\begingroup\small
\begin{tabular}{llllllll}
\toprule
Christianity & Tech Gadgets & Sports & TV or Show & Social Media & Wildfire & Book or Movie & Problem \\
\midrule
life & apple & game & show & post & area & story & may \\
god &  feature &  season &  live &  share &  water &  book &  change \\
word &  phone &  play &  night &  free &  fire &  movie &  number \\
heart &  device &  team &  star &  comment &  north &  film &  deal \\
church &  user &  against &  news &  video &  land &  full &  result \\
pope &  plus &  player &  series &  photo &  west &  character &  allow \\
father &  update &  football &  special &  click &  near &  author &  note \\
christian &  version &  yard &  tv &  link &  south &  title &  problem \\
son &  iphone &  coach &  award &  facebook &  california &  director &  within \\
lord &  app &  ball &  fan &  twitter &  local &  writer &  step \\
\bottomrule
 &  &  &  &  &  &  &  \\
\toprule
Law, Police & Today & Event Date & Design & Reports & Event Location & Help & Systems \& Security \\
\midrule
case & new & \_year\_ & include & report & city & use & service \\
 law &  best &  september &  add &  plan &  event &  need &  system \\
 police &  today &  th &  design &  issue &  center &  help &  data \\
 court &  next &  watch &  create &  member &  st &  different &  technology \\
 claim &  open &  online &  large &  public &  street &  small &  customer \\
 charge &  month &  date &  image &  accord &  art &  easy &  network \\
 officer &  late &  october &  view &  continue &  park &  save &  access \\
 against &  hour &  august &  base &  national &  friday &  important &  solution \\
 act &  york &  episode &  space &  action &  sept &  type &  security \\
 judge &  check &  july &  form &  official &  monday &  choose &  provide \\
\bottomrule
\end{tabular}
\endgroup
\end{table*}

%% file: 7_conclusion.tex
\section{Conclusion}
\label{conclusion}
While the structure of DAPPER mimics its predecessor, we derive a fundamentally new inference algorithm based on CVI. DAPPER surpasses its predecessor in terms of speed (35x faster), memory (constant requirements for mini-batch training), and significantly better likelihoods. DAPPER scales to massive datasets on commodity hardware, which in turn allows for deeper insights into topics, and common narratives hidden in the data.  Additionally, we show that Regularized Variational Inference, which is applied to the DAPPER model to encourage distrinct personas, integrates with CVI cleanly because CVI preserves closed form updates. The success of DAPPER demonstrates that CVI can be applied to complex, temporal graphical models --- eliminating the need to run multiple optimization procedures on each document, and instead replace all parameter updates with fast, closed form updates and stochastic mini-batch training.

While the work presented here demonstrates the DAPPER topic model's readiness for industrial-sized problems, there exist opportunities for further research. For one, our results show that too noisy of updates resulting from small mini-batches lead to poor performance. However, as briefly mentioned in the results, simple attempts to reduce variance through gradient averaging did not yield performance improvements. Further research is needed to find gradient updates that improve model performance in early iterations (as with small to medium batch sizes), but converge to better PWLLs (as with the larger batch sizes). Additionally, the DAPPER model requires time to be discretized in the data, and while the variational Kalman Filter keeps results from being too sensitive to the window size chosen, new methods exist capable of discretizing time based on shifts in topics \cite{chen2018incremental}.

%% file: 8_acknowledgements.tex
\section*{Acknowledgments}
We thank reviewers for their valuable comments, University of Minnesota Supercomputing Institute (MSI) for technical support, and CaringBridge for their support and collaboration. The research was supported by NSF grants IIS-1563950, IIS-1447566, IIS-1447574, IIS-1422557, CCF-1451986, CNS-1314560.